\documentclass[conference]{aiaa-pretty}
\usepackage[utf8]{inputenc}
\usepackage[english]{babel}
\usepackage{graphicx}
\usepackage{caption}
\usepackage{listings}
\usepackage[linesnumbered,ruled,vlined]{algorithm2e}

\usepackage{xcolor}

\author[Jerry]{ %
Jun Xiang\thanks{Ph.D. Student, Department of Aerospace Engineering, AIAA Student Member, \texttt{jxiang9143@sdsu.edu }},
Drake Essick%
    \thanks{Undergraduate student, Department of Electrical and Computer Engineering, \texttt{dessick3793@sdsu.edu} }
 , 
 Luiz Gonzalez Bautista%
    \thanks{Undergraduate student, Department of Electrical and Computer Engineering, \texttt{lgonzalezbauti4071@sdsu.edu }},
Junfei Xie \thanks{Associate Professor, Department of Electrical and Computer Engineering, AIAA Member, \texttt{jxie4@sdsu.edu}},
Jun Chen\thanks{Assistant Professor, Department of Aerospace Engineering, AIAA Senior Member, \texttt{Jun.Chen@sdsu.edu}}\\
\textit{San Diego State University, San Diego, CA 92182}}

\title{Landing Trajectory Prediction for UAS Based on Generative Adversarial Network}

 \abstract{

Models for trajectory prediction are an essential component of many advanced air mobility studies. These models help aircraft detect conflict and plan avoidance maneuvers, which is especially important in Unmanned Aircraft systems (UAS) landing management due to the congested airspace near vertiports.
In this paper, we propose a landing trajectory prediction model for UAS based on Generative Adversarial Network (GAN). The GAN is a prestigious neural network that has been developed for many years. In previous research, GAN has achieved many state-of-the-art results in many generation tasks. The GAN consists of one neural network generator and a neural network discriminator. Because of the learning capacity of the neural networks, the generator is capable to understand the features of the sample trajectory. The generator takes the previous trajectory as input and outputs some random status of a flight. According to the results of the experiences, the proposed model can output more accurate predictions than the baseline method(GMR) in various datasets. To evaluate the proposed model, we also create a real UAV landing dataset that includes more than 2600 trajectories of drone control manually by real pilots.

}

\begin{document}

\maketitle

\section{Introduction}

Unmanned aerial vehicles (UAVs) are the future of aviation. FAA forecasts that there will be 1.81 million units of drone fleet by 2026~\cite{FAA}. With so many UAVs flying in the airspace, tracking and predicting the motion of UAVs become important. The security service needs to predict misused drones in public areas including airports, mass events, or public demonstrations~\cite{laurenzis2020prediction}. Meanwhile, in-network UAVs need to predict the trajectory of out-network UAVs to avoid them, which is especially important in UAV landing management due to the congested airspace near vertiports~\cite{zhou2020optimized}. Conflict detection and resolution(CD\&R) technology, plays an important role in the air traffic system~\cite{lin2018uav,wu2022safety} and highly relies on accurate
and timely predictions of out-network UAV trajectories~\cite{kang2020model}. There are many methods have been proposed to predict the trajectory of UAVs.  
However, existing methods for UAV trajectory prediction are ordinarily very slow and unable to be applied online~\cite{xie2022efficient}. In addition to fast prediction, higher accuracy of prediction results is required. Many planning methods are bothered by the uncertainty of the future prediction.

Aircraft trajectory prediction is always a popular research topic and there are many methods proposed. The Recurrent Neural network is proposed to predict aircraft trajectory with weather features\cite{pang2019recurrent}. The Hidden Markov Model (HMM) is recently applied to predict trajectories considering environmental uncertainties\cite{ayhan2016aircraft}.  The deep generative
convolutional recurrent neural network (RNN) approach is applied for 4D trajectory prediction\cite{liu2018predicting}. A hybrid 3-dimensional UAV modeling framework, which integrates the physics-based and data-based models, can predict the motion of UAVs using only a small amount of real fight data\cite{wang20193}. Some aircraft prediction methods focus on modeling aircraft trajectories around the terminal. One traditional way is to predict the aircraft near the terminal by the Physical equation of motion\cite{chatterji1999short}. The bayesian network can learn
statistical representations of dynamic variables of aircraft and predict the motion\cite{kochenderfer2010airspace}. Probabilistic generative model\cite{barratt2018learning} predicts the aircraft trajectory by learning from position measurements. 

The neural network generator has been proven to predict many objects such as language, image, and trajectories of pedestrians. Transformer~\cite{vaswani2017attention} is the state-of-the-art method to draw global dependencies between input and output. BERT~\cite{devlin2018bert}, uses Transformer as one of the encoders and is able to understand and process natural language. GATs~\cite{velivckovic2017graph} achieved the state-of-the-art result across four established transductive and inductive graph benchmarks leveraging masked self-attentional layers. At the same time, the transformer has successfully been used with GANs. SAGAN~\cite{zhang2019self} achieved pioneering image generation results by using a transformer layer in GANs. ViTGAN~\cite{lee2021vitgan}achieves a comparable performance of image generation to state-of-the-art CNN-based GANs without using convolution or pooling. CNN also performs decently recognizing image and text~\cite{long2015fully,ren2015faster} \cite{he2016deep,huang2017densely}. GANs have proven to be able to capture the movement of mobile objects and predict their trajectories in many existing studies. For example, 
Social-GAN~\cite{gupta2018social} is able to predict pedestrians' movement considering the social interaction between pedestrians. SoPhie~\cite{sadeghian2019sophie} can understand the scene context besides social behavior and generate pedestrian trajectories.  CGAN was proposed in~\cite{pang2020conditional} to solve weather-related aircraft trajectory prediction problems. 3D-GAN~\cite{wu2016learning} can generate high-quality 3D objects without supervision. Therefore it is rational to train a Neural network to predict the landing trajectory of the UAS.  


The paper proposes to  predict the landing trajectory for UAS with a GAN-based method. The major contributions of this paper are summarized as follows:
\begin{itemize}
    \item 
    This paper adopts a 2D trajectory GAN-based method to predict 3D landing trajectories of UAS. The proposed method can generate prediction very fast and outperform the Gaussian Mixture Regression(GMR) in accuracy.
    \item 
    This paper provides a real UAS landing trajectory dataset for future research.  
    \item
    After GANs are well-trained, the discriminator part can be used to evaluate other predictions.

\end{itemize}

The paper's structure is as follows. Section~\ref{sec:background} explains the background of GANs and GMM; Section \ref{sec: method} introduces the trajectory, how the dataset is created, and the architecture of the generator; Section~\ref{sec: results} shows the results by comparing the GANs and GMM; Section~\ref{sec: conc} concludes this paper.


\section{Background}\label{sec:background}
\subsection{LSTM Generative Adversarial Network}
Long Short-Term Memory (LSTM)~\cite{hochreiter1997long} is a novel recurrent network architecture in conjunction with an appropriate gradient-based learning algorithm. The LSTM can enforce constant error flow so it can accelerate learning store information over extended time intervals. LSTM consists of multiplicative gate units that can learn to open or close the access to the error flow. LSTM can be used as the encoder and the decoder in GANs.

A Generative Adversarial Network consists of two neural networks, generator and discriminator, trained in opposition to each other~\cite{goodfellow2014generative}. The training procedure is similar to a two-player min-max game with the following objective function:
\begin{equation}
    \min\limits_{G} \max\limits_{D} V(D,G) = \mathbb{E}_{x\sim p_{data}(x)}[\log{D(x)}] + \mathbb{E}_{z\sim p_{z}(z)}[\log{1-D(G(z))}]
\end{equation}
In this paper, G denotes the generator that inputs factors and generates trajectory. z denoted the input factors that decide how a flight should fly. Generator G takes the input factors and output trajectory G(z). D denotes discriminator and D(x) output the probability that x is a trajectory is desired. 

 Gupta proposed social-GAN~\cite{gupta2018social} that can understand human motion behavior. The social-GAN is able to observe the motion trajectories of pedestrians and predict the trajectories of pedestrians. Social-GAN combines tools from sequence prediction and generative adversarial networks. The generator of the social-GAN is consists of an LSTM-based encoder, a pooling module, and an LSTM-based decoder. The LSTM encoder encodes the motion trajectories to hidden items $h$. Then the pooling module pool the hidden items $h$ while combining desired information $i$ and the hidden items $h$ together. Finally, the decoder decodes the hidden items and outputs the predicted trajectory.  

\subsection{Gaussian Mixture Regression predictor}
In this paper, Gaussian Mixture Regression(GMR) predictor is the baseline method. GMR predictor was proposed in \cite{calinon2007learning} and the GMR predictor we use in this paper is proposed in\cite{fabisch2021gmr}. The GMR predictor is one of the most popular predictors. The training process of GMR is shown as equation~(\ref{gmrt}).

\begin{equation}\label{gmrt}
p(\boldsymbol{x}, \boldsymbol{y})=\sum_{k=1}^K \pi_k \mathcal{N}_k\left(\boldsymbol{x}, \boldsymbol{y} \mid \boldsymbol{\mu}_{\boldsymbol{x} \boldsymbol{y}_k}, \boldsymbol{\Sigma}_{\boldsymbol{x} \boldsymbol{y}_k}\right)
\end{equation}
where $\mathcal{N}_k\left(\boldsymbol{x}, \boldsymbol{y} \mid \boldsymbol{\mu}_{\boldsymbol{x} \boldsymbol{y}_k}, \boldsymbol{\Sigma}_{\boldsymbol{x} \boldsymbol{y}_k}\right)$ are Gaussian distributions with mean $\boldsymbol{\mu}_{\boldsymbol{x} \boldsymbol{y}_k}$ and covariance $\boldsymbol{\Sigma}_{\boldsymbol{x}\boldsymbol{y}_k}$, $K$ is the number of Gaussians, and $\pi_k \in [0, 1]$ are priors that sum up to one. 

After the GMR is trained, GMR predicts distributions of variables $\boldsymbol{y}$ by computing the conditional distribution $p(\boldsymbol{y} \mid \boldsymbol{x})$. The conditional distribution of each individual Gaussian
\begin{equation}
\mathcal{N}\left(\boldsymbol{x}, \boldsymbol{y} \mid \boldsymbol{\mu}_{x y}, \boldsymbol{\Sigma}_{x y}\right) \\
\boldsymbol{\mu}_{\boldsymbol{x} y}=\left(\begin{array}{c}
\mu_x \\
\mu_y
\end{array}\right), \quad \boldsymbol{\Sigma}_{x y}=\left(\begin{array}{cc}
\boldsymbol{\Sigma}_{x \boldsymbol{x}} & \boldsymbol{\Sigma}_{\boldsymbol{x} y} \\
\boldsymbol{\Sigma}_{y x} & \boldsymbol{\Sigma}_{y y}
\end{array}\right)
\end{equation}

is defined by

\begin{equation}
\mu_{y \mid x}=\mu_y+\Sigma_{\boldsymbol{y} x} \Sigma_{x \boldsymbol{x}}^{-1}\left(x-\mu_x\right) \\
\end{equation}
\begin{equation}
\Sigma_{y \mid x}=\Sigma_{y y}-\Sigma_{y x} \Sigma_{x \boldsymbol{x}}^{-1} \Sigma_{x y}
\end{equation}

 the conditional distribution of each individual Gaussian and their priors is computed by equation
\begin{equation}
\pi_{\boldsymbol{y} \mid \boldsymbol{x}_k}=\frac{\mathcal{N}_k\left(\boldsymbol{x} \mid \boldsymbol{\mu}_{\boldsymbol{x} k}, \boldsymbol{\Sigma}_{\boldsymbol{x} k}\right)}{\sum_{l=1}^K \mathcal{N}_l\left(\boldsymbol{x} \mid \boldsymbol{\mu}_{\boldsymbol{x} l}, \boldsymbol{\Sigma}_{\boldsymbol{x} l}\right)}
\end{equation}

to obtain the conditional distribution
\begin{equation}
p(\boldsymbol{y} \mid \boldsymbol{x})=\sum_{k=1}^K \pi_{\boldsymbol{y} \mid \boldsymbol{x}_k} \mathcal{N}_k\left(\boldsymbol{y} \mid \boldsymbol{\mu}_{\boldsymbol{y} \mid \boldsymbol{x}_k}, \boldsymbol{\Sigma}_{\boldsymbol{y} \mid \boldsymbol{x}_k}\right)
\end{equation}

\section{Method}\label{sec: method}
\subsection{Trajectory}
Mathematically, Let T denotes the trajectory sequence of consists of multiple flight statuses. Let $x_{t+1} \in R^n$ the flight status at $t + 1$,  will be determined by the previous status, dynamic model, and control input according to ordinary differential equation (ODE):
\begin{equation}
    T_{t_1, t_2} = [x_{t_1}, x_{t_1+1}, ...x_{t_2 - 1} ,x_{t_2}]
\end{equation}

\begin{equation}
    x_{t+1} = f(x_{t}, a_t(E)), a_t \in \mathbb{A}
\end{equation}
where $f$ denotes flight the dynamic model. $a_t$, denotes the control inputs which how the pilot reacts the environment E at time t. Let $a(E)$ called action function in these paper. $\mathbb{A}$ denotes the set of all the possible control inputs. However, in real world, flight dynamic model has noises and high computational cost. On the other hands, environment factors and previous status are only variables required to derive a trajectory. In this paper, the only flight status we focus on is position.

In order to predict the future trajectory without flight dynamic model and action function, the generator model(G) was proposed:
\begin{equation}
    T_{t_1, t_2} = G(T_{t_0,t_1})
\end{equation}
where the Generator takes the previous trajectory sequence as input and outputs a predicted trajectory sequence. In this paper, all the previous trajectory sequence contains 10 points, and all the predicted trajectory sequence contains 10 points as well.

\subsection{Data Collection}\label{data}
In this paper, we prepare three different types of datasets to train the proposed model and test the learning capacity. 

\begin{figure}
    \centering
    \includegraphics[width = 0.8\textwidth]{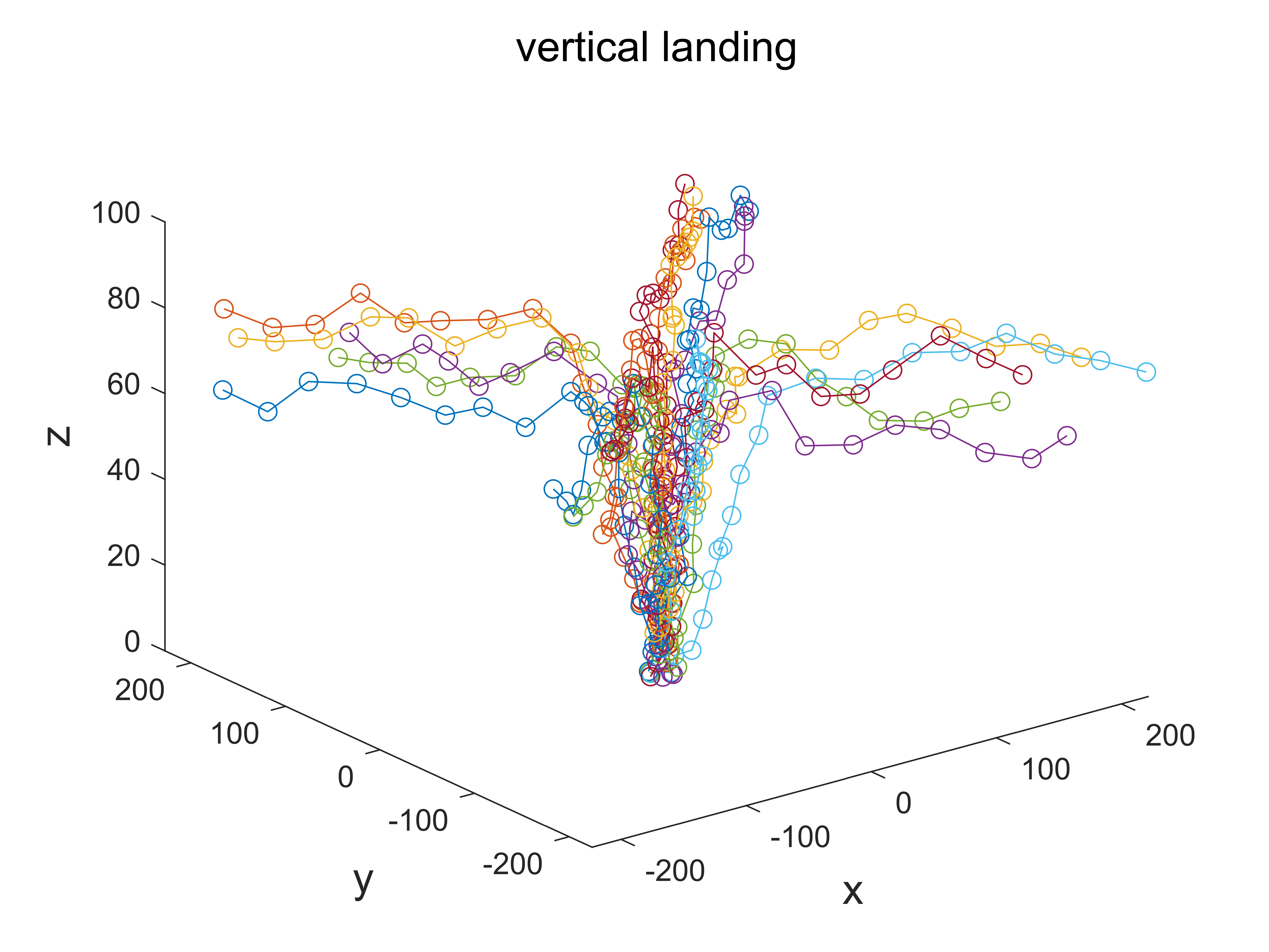}
    
    \caption{Vertical Landing Trajectory}
    \label{vertical}
\end{figure}
\subsubsection{Vertical Landing Data}
We will start with a sample dataset as shown in Figure~\ref{vertical} to test how the generator works. The dataset includes the trajectories starting from four different directions to the destination (0,0,0). The average initial position is (-200, 200, 75) for direction1, (200, -200, 75) for direction2, (-200, -200, 75) for direction3, and (200, 200, 75) for direction4. The variance for the x, and y positions is 50, and for the z position is 3.5. The drone first flies horizontally to the points right above the destination, then land vertically.

\begin{figure}
    \centering
    \includegraphics[width = 0.8\textwidth]{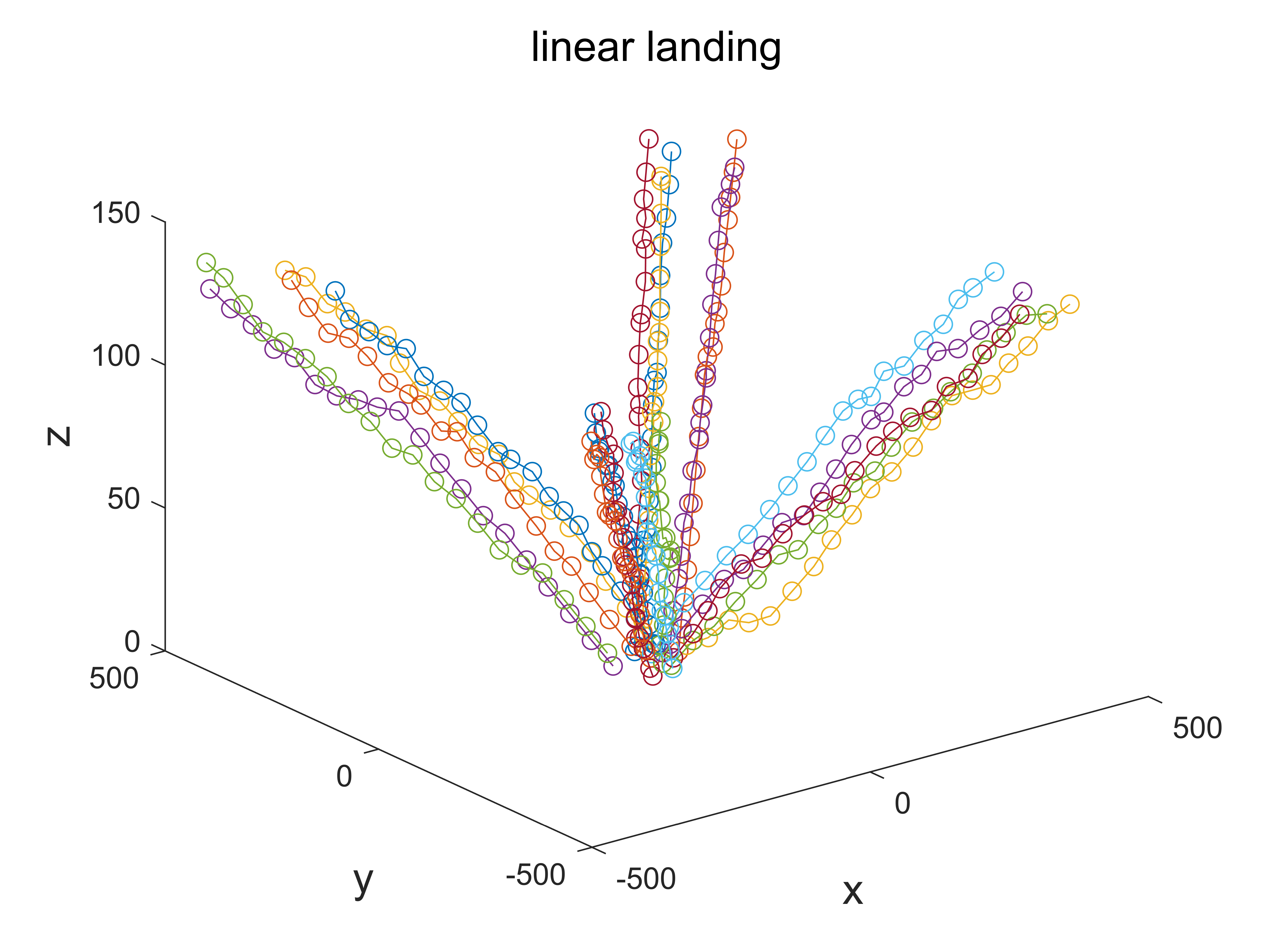}
    
    \caption{Linear Landing Trajectory}
    \label{linear}
\end{figure}

\subsubsection{Linear Landing Data}
The second dataset (as shown in Figure~\ref{linear}) we used to evaluate this method is simulated linear landing data. Same as vertical data, the drone flies from four different initial positions and land at position (0,0,0). The average initial position is (-400, 400, 140) for direction1, (400, -400, 140) for direction2, (-400, -400, 140) for direction3, and (400, 400, 140) for direction4. The variance for the x, and y positions is 100, and for the z position is 7.5. In contrast to vertical landing, the drone flight flies straight forward to the destination.

\subsubsection{Real Data}
\begin{figure}
    \centering
    \includegraphics[width = 0.8\textwidth]{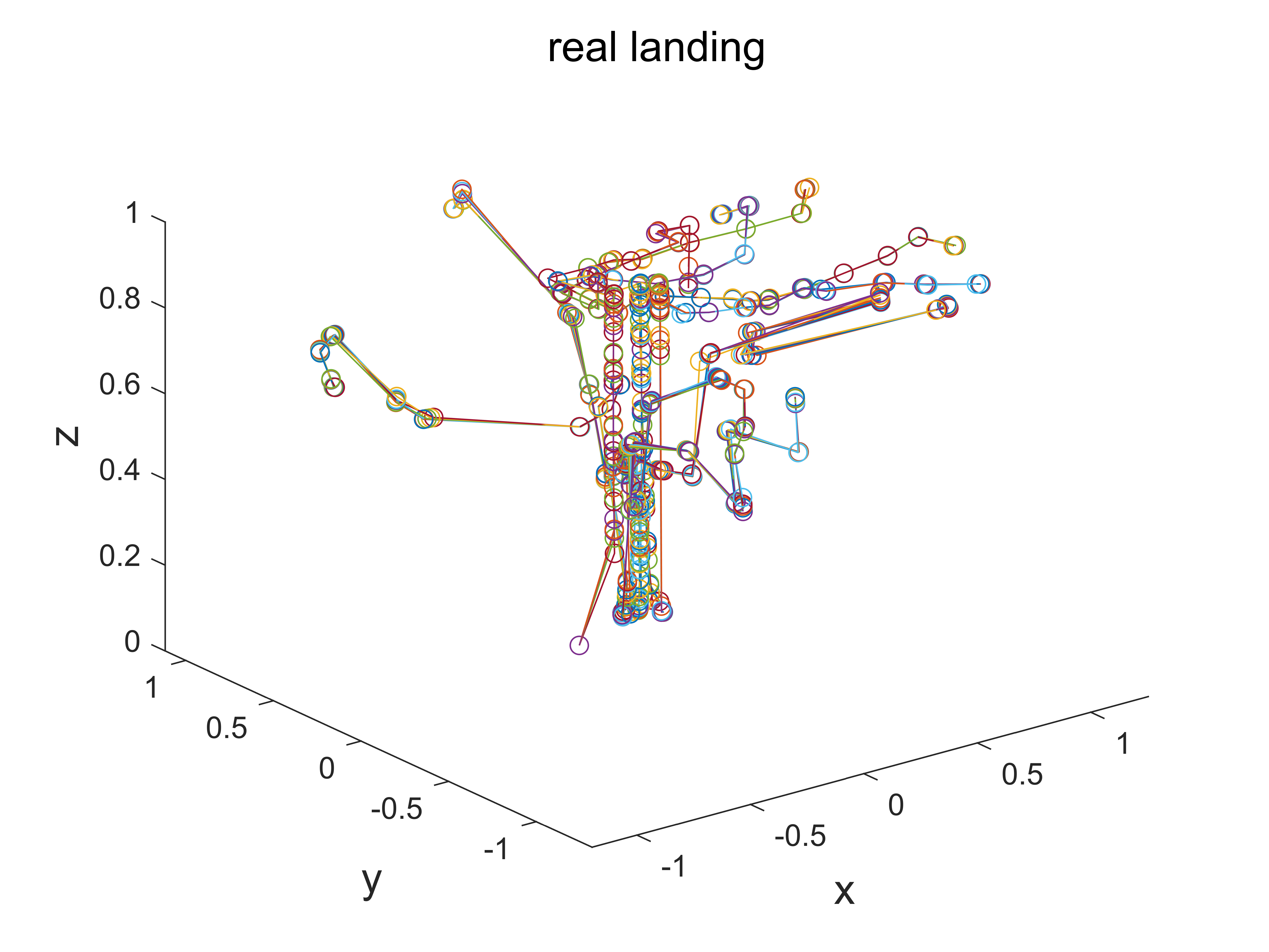}
    
    \caption{Real Landing Trajectory}
    \label{real}
\end{figure}

\begin{figure}
    \centering
    \includegraphics[width = 0.45\textwidth]{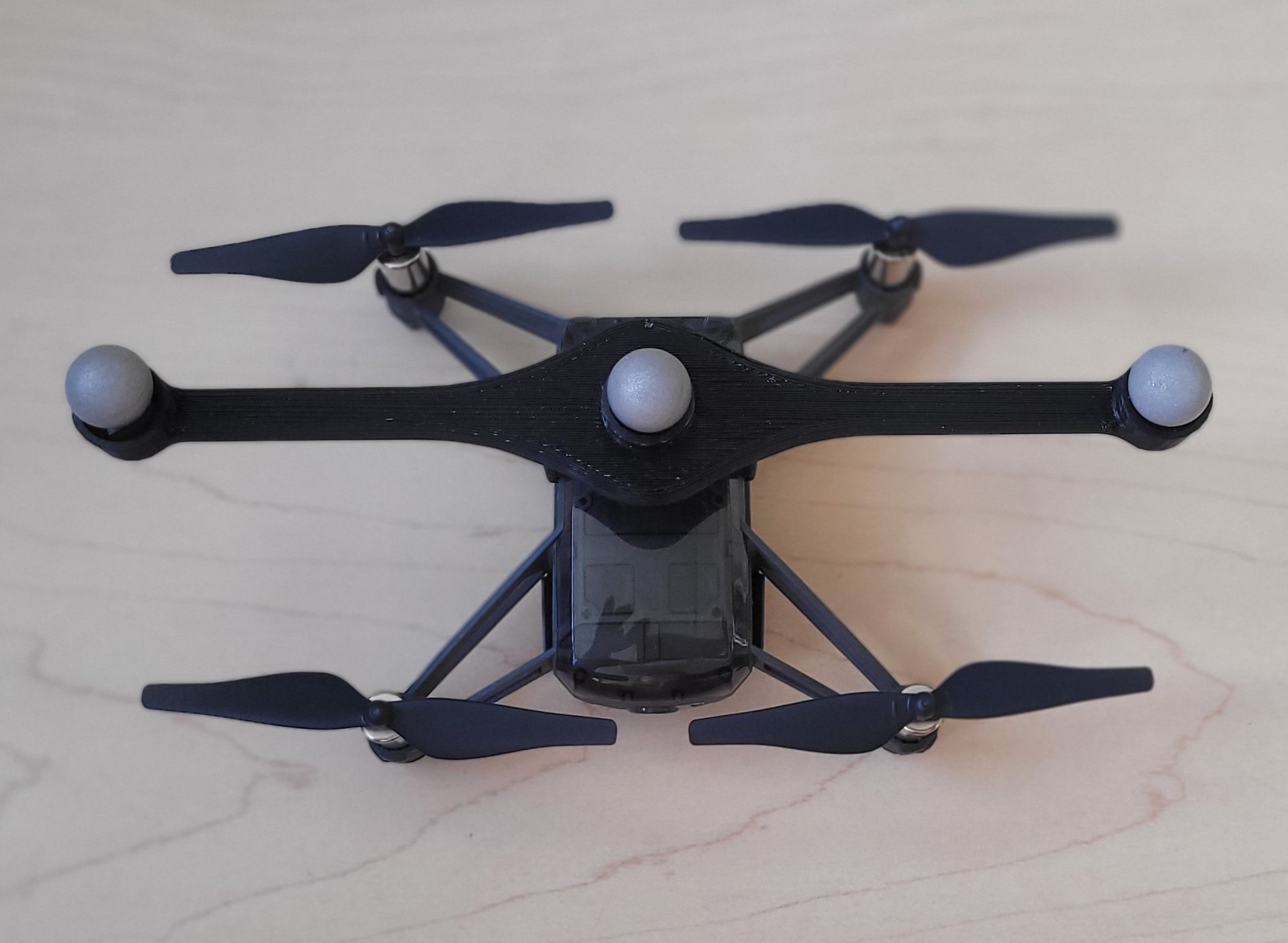}
    \includegraphics[width = 0.45\textwidth]{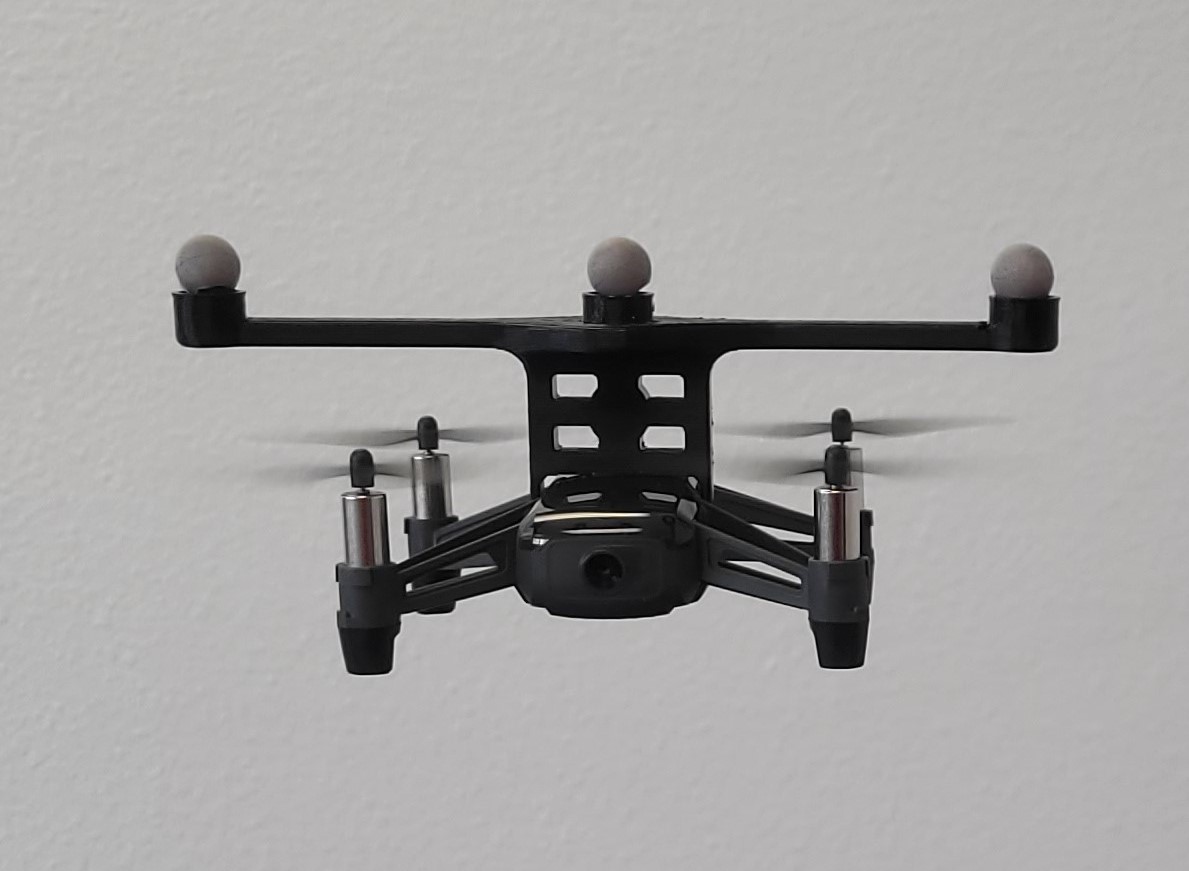}

    \caption{DJI Tello Drone}
    \label{drone}
\end{figure}

Real trajectories as shown in Figure~\ref{real} will be collected by the real pilot. Similar to the simulated data, the real pilot flies the drone from different initial positions and land at the destination. The initial position and trajectory pattern will fully rely on how the real pilot wants to control the drone. In this paper, the drone, as shown in Figure~\ref{drone}, is a customized DJI Tello, which we can track and localize by the motion capture system.

\subsection{LSTM-Pooling based Generator}
The trajectory can be treated as a kind of data sequence, therefore, LSTM, which is one of the best neural network structures to understand data sequences, can be used to understand and generate the trajectory. Social-GANs\cite{gupta2018social} was proposed to predict pedestrians' future trajectory based on their past trajectory and social interaction. Similar to social-GANs, We show one of the trajectories that are desired by the LSTM and tell LSTM some factors, then the LSTM generates trajectories based on the trajectory we showed to it. 

The sequence data are needed to be embedded into a fixed-length vector before sending it into the LSTM cell. The LSTM cell will encode the embedded vector and previous hidden item into a current hidden item.

\begin{align}
&e_t = \sigma(x_t) \\
& h^{el}_t = LSTM_{encoder}(h^{el}_{t-1}, e_t; W_{encoder})
\end{align}
where $e_t$ denotes the embedding vector. $x_t$ is one of the flight statuses in the trajectory sequence. $\sigma$ is the preprocessing function.  $h_t$ is the hidden time for time $t$. The $W_{LSTM}$ is the LSTM weight. The pooling will input the environment factors and combine them into hidden items come from the LSTM encoder layer:

\begin{equation}
    p = P(E, z)
\end{equation}
\begin{equation}
    e_t = \phi(x_{t-1};W_{em})
\end{equation}
The the LSTM decoder layer will take the combined hidden item, generate the trajectory.
\begin{equation}
    h^{dl}_{t} = LSTM_{decoder}(\gamma(h^{dl}_{t-1},p), e_t;W_{decoder})
\end{equation}
\begin{equation}
    x_t = \gamma(h^{dl}_t)
\end{equation}

\begin{figure}
    \centering
    \includegraphics[width = 0.8\textwidth]{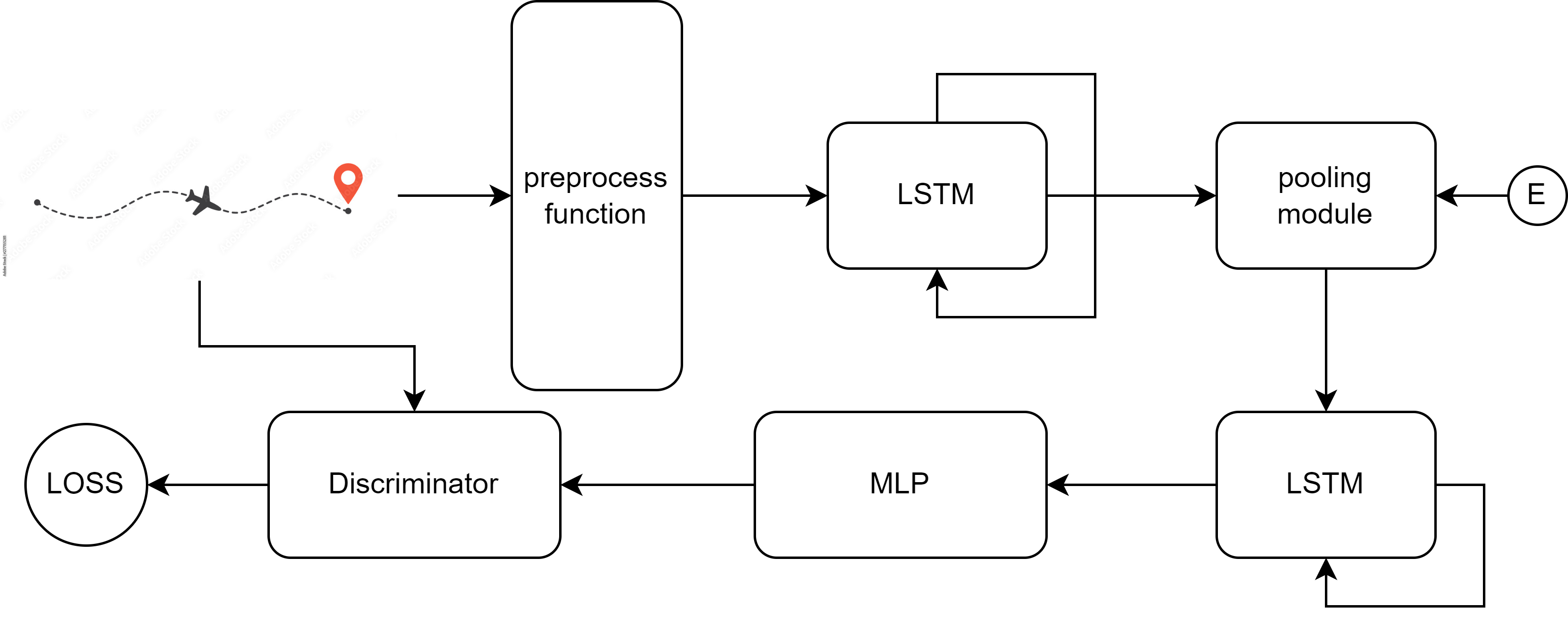}
    \caption{Training Process}
    \label{TP}
\end{figure}

\subsection{Discriminator}
Instead of a traditional mathematical model, we use a neural network as a discriminator to evaluate the generated trajectory. Based on the author's knowledge, there is no traditional mathematical model that is completely able to identify if a trajectory is a good trajectory. The discriminator inputs the trajectory T and gives a score s. If the given data is real, the discriminator is expected to give a high score. The score s is the main component of the loss function of the generator. In this paper, the discriminator is an LSTM network.
\begin{equation}
    s = LSTM(T)
\end{equation}

\section{Result}\label{sec: results}
\begin{figure}
    \centering
    \includegraphics[scale = 0.4]{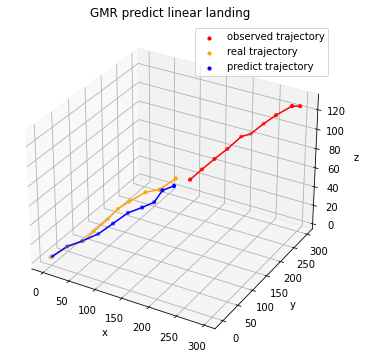}
        \includegraphics[scale = 0.4]{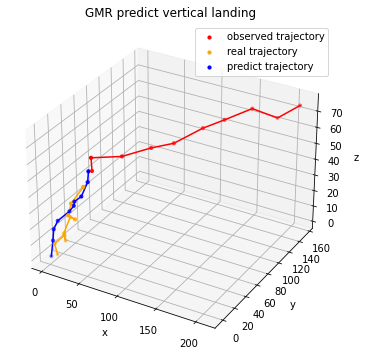}
            \includegraphics[scale = 0.4]{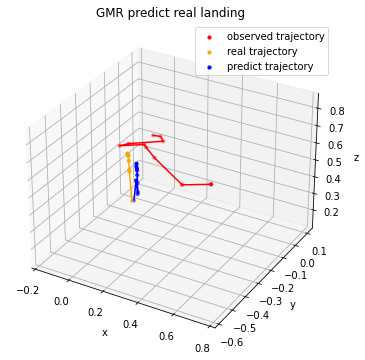}
    
    \caption{GMM prediction}
    \label{GMR}
\end{figure}
\begin{figure}
    \centering
    \includegraphics[scale = 0.4]{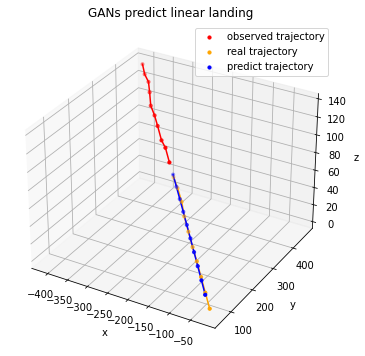}
        \includegraphics[scale = 0.4]{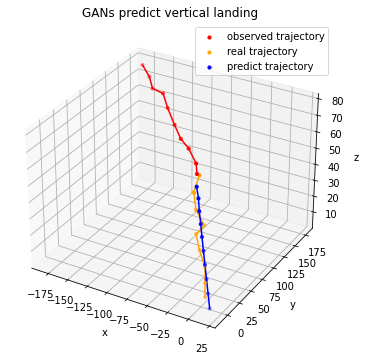}
            \includegraphics[scale = 0.4]{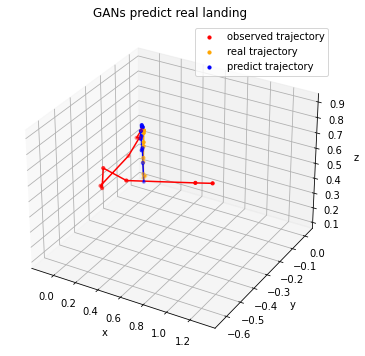}

    \caption{GANs prediction}
    \label{GANs}
\end{figure}
\begin{figure}
    \centering
    \includegraphics[scale = 0.8]{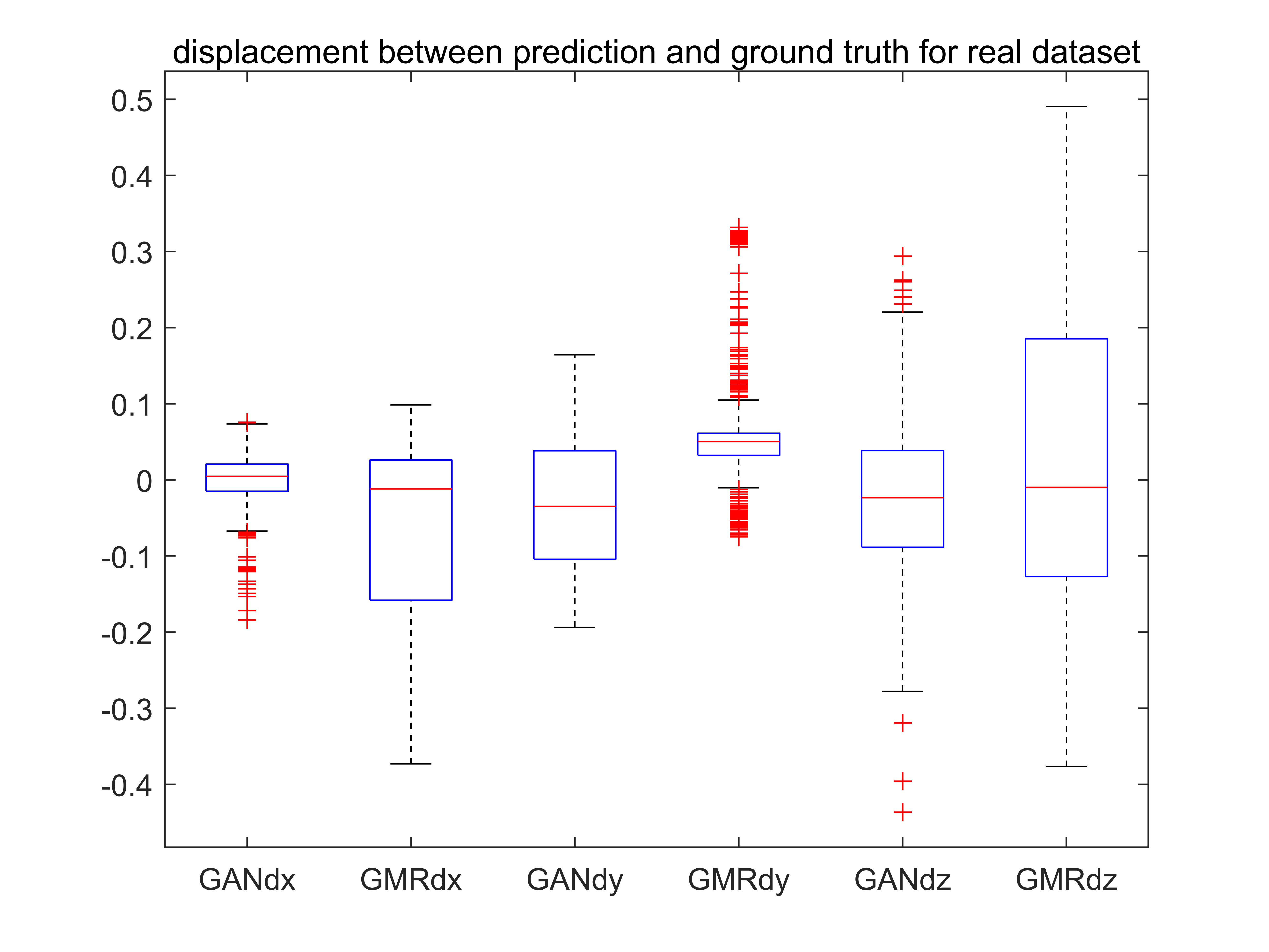}

    \caption{prediction displacement in each axis(real dataset)}
    \label{displacement}
\end{figure}
The proposed GANs will be tested in three prepared datasets introduced in section~\ref{data}.
For each dataset, there will be 3000(2430 for real) trajectories for training and 100 trajectories for evaluation. Figure~\ref{GMR},\ref{GANs} shows examples of the GMR and GAN predicting the landing trajectory. The accuracy of the prediction will be evaluated by the Average Displacement Error(ade) between the predicted trajectory and the real trajectory. Then, we will test if the discriminator can distinguish good and bad predictions.

\subsection{Average Displacement Error performance}
The equation to calculate ade for one point is shown as equation~\ref{ade}
\begin{equation}\label{ade}
    ade = mean(\sqrt{(dx)^2 + (dy)^2 + (dz)^2})
\end{equation}
where dx is the displacement between the predicted point and the ground truth point in the x-axis, dy is the displacement in the y-axis and dz is the displacement in the z-axis. 
The trained GAN and GMR observe 100 evaluation trajectories and predict possible trajectories. The ade is the mean of 100 displacement errors between prediction and ground truth. The ade result for each point and the dataset is shown as tabel~\ref{tab: ade}. The results show that GAN has a very good performance to predict the first four points, while the predicts get worse as the trajectory gets longer. For the vertical and linear datasets, GAN only outperforms GMR for the first four points. However, for the real dataset, GAN outperforms GMR for all the points. Because the vertical and linear dataset is simulated and has a smaller variance, it is easier for GMR, which is a Gaussian-based method, to predict. In contrast, in the real dataset, real pilots control the drone more randomly and the control logic is more subtle, GAN is able to mine the hidden rules and gets better results. 
Figure~\ref{displacement} shows a comparison of the prediction displacement in each axis for the real dataset between GAN and GMR. The displacement of GMR predictions obviously larger than GAN in all axes. In conclusion, GAN has much better accuracy in predicting the trajectories from the real dataset, while performing decently in predicting the trajectories from the simulated dataset. However, the longer the predicting is, the worse the GAN will perform. 
\begin{table}[hbt!]
\caption{\label{tab: ade} Average Displacement Error for each point(lower is better)}
\centering
\begin{tabular}{c c c c c c c}
\hline
\hline
 & \multicolumn{2}{c}{vertical} & \multicolumn{2}{c}{linear}   &  \multicolumn{2}{c}{real}  \\ \hline
method& GAN & GMR & GAN & GMR & GAN & GMR\\
\hline
point1 &7.13$\pm$2.13 & 8.61$\pm$3.53 & 5.01$\pm$2.25 & 6.17$\pm$2.51 & 0.10$\pm$0.05&0.15$\pm$0.06\\
point2 &9.11$\pm$4.2 & 11.12$\pm$4.87 & 7.31$\pm$3.44 & 9.1 $\pm$3.86 & 0.11$\pm$0.04&0.18$\pm$0.07\\
point3 &10.83$\pm$4.96 & 12.49$\pm$5.52 & 8.93$\pm$4.55 & 10.05$\pm$4.5 & 0.11$\pm$0.03&0.22$\pm$0.09\\
point4 &12.11$\pm$5.56 & 14.04$\pm$5.73 & 10.4$\pm$5.33 & 11.28$\pm$5.19 & 0.12$\pm$0.03&0.23$\pm$0.12\\
point5 &12.99$\pm$5.92 & 14.3$\pm$5.81 & 11.88$\pm$6.12 & 11.72$\pm$5.25 & 0.13$\pm$0.04&0.23$\pm$0.09\\
point6 &13.44$\pm$6.32 & 14.19$\pm$5.67 & 12.67$\pm$6.4 & 12.57$\pm$6.13 & 0.14$\pm$0.04&0.23$\pm$0.11\\
point7 &14.79$\pm$6.76 & 13.99$\pm$5.69 & 14.07$\pm$6.51 & 12.45$\pm$5.69 & 0.15$\pm$0.05&0.23$\pm$0.1\\
point8 &15.06$\pm$7.09 & 12.52$\pm$5.81 & 14.97$\pm$7.68 & 12.27$\pm$5.41 & 0.14$\pm$0.06&0.24$\pm$0.1\\
point9 &15.02$\pm$6.97 & 10.92$\pm$4.07 & 16.04$\pm$8.53 & 11.46$\pm$5.62 & 0.14$\pm$0.07&0.21$\pm$0.16\\
point10 &14.88$\pm$8.41 & 9.0$\pm$3.69 & 17.01$\pm$8.97 & 10.93$\pm$5.72 & 0.18$\pm$0.08&0.24$\pm$0.21\\
\hline
\hline
\end{tabular}
\end{table}

\subsection{Discriminator}
The discriminator of the GANs can evaluate a trajectory and give a score. We show the discriminator 100 true trajectories and 100 fake trajectories from each dataset and let the discriminator generates the scores for those trajectories. The results are shown as table~\ref{tab: score}. According to the result, the discriminator gives higher scores to the true trajectories. It proves that the discriminator can distinguish true and fake trajectories.
\begin{table}[hbt!]
\caption{\label{tab: score} score given to real and fake trajectories by the discriminator}
\centering
\begin{tabular}{c c c c c c c}
\hline
\hline
 & \multicolumn{2}{c}{vertical} & \multicolumn{2}{c}{linear}   &  \multicolumn{2}{c}{real}  \\ \hline
method& True & Fake & True & Fake & True & Fake\\
\hline
score &2.79$\pm$0.31 & -9.7$\pm$0.43 & 2.99$\pm$0.77 & -7.93$\pm$0.67 & 2.32$\pm$2.76&0.21$\pm$0.86\\
\hline
\hline
\end{tabular}
\end{table}

\section{Conclusion}\label{sec: conc}
In this paper, the LSTM-GAN-based method can accurately predict the motion of UAVs during the landing. The prediction error is lower than the baseline method. Meanwhile, the discriminator can give a high score to ground truth trajectories while giving a low score to fake trajectories. It proves that the trained discriminator can be used to evaluate the prediction. A dataset that includes a large number of trajectories of UAVs controlled by real pilots is also created. 
However, according to the results, the accuracy of prediction results becomes lower as the prediction length is longer. This method may not work for long-term prediction. In the future, a new neural network encoder and decoder, which can process a larger amount of information is required for this GANs framework.

\bibliographystyle{aiaa}
\bibliography{main}

\end{document}